\documentclass[10pt,twocolumn,letterpaper]{article}

\usepackage{iccv}
\usepackage{times}
\usepackage{epsfig}
\usepackage{graphicx}
\usepackage{amsmath}
\usepackage{amssymb}
\usepackage{multirow}
\usepackage{array}

\usepackage[pagebackref=true,breaklinks=true,colorlinks,bookmarks=false]{hyperref}

\iccvfinalcopy 

\def\iccvPaperID{****} 

\ificcvfinal\fi
\allowdisplaybreaks

\begin{document}
\hbadness=1000000000
\vbadness=1000000000
\hfuzz=10pt
\setlength{\abovedisplayskip}{6pt}
\setlength{\belowdisplayskip}{6pt}
\setlength{\textfloatsep}{10pt plus 1.0pt minus 2.0pt}

\title{Partial Video Domain Adaptation with \\Partial Adversarial Temporal Attentive Network}

\author{Yuecong Xu, Jianfei Yang, Haozhi Cao, Qi Li, Kezhi Mao\\
School of Electrical and Electronic Engineering, Nanyang Technological University, Singapore\\
50 Nanyang Avenue, Singapore 639798\\
{\tt\small \{xuyu0014, yang0478, haozhi001, liqi0024\}@e.ntu.edu.sg, ekzmao@ntu.edu.sg}
\and
Zhenghua Chen\\
Institute for Infocomm Research, Agency for Science, Technology and Research (A*STAR), Singapore\\
1 Fusionopolis Way, \#21-01, Connexis South, Singapore 138632\\
{\tt\small chen-zhenghua@i2r.a-star.edu.sg}
}

\maketitle
\ificcvfinal\thispagestyle{empty}\fi

\begin{abstract}
   Partial Domain Adaptation (PDA) is a practical and general domain adaptation scenario, which relaxes the fully shared label space assumption such that the source label space subsumes the target one. The key challenge of PDA is the issue of negative transfer caused by source-only classes. For videos, such negative transfer could be triggered by both spatial and temporal features, which leads to a more challenging Partial Video Domain Adaptation (PVDA) problem. In this paper, we propose a novel Partial Adversarial Temporal Attentive Network (PATAN) to address the PVDA problem by utilizing both spatial and temporal features for filtering source-only classes. Besides, PATAN constructs effective overall temporal features by attending to local temporal features that contribute more toward the class filtration process. We further introduce new benchmarks to facilitate research on PVDA problems, covering a wide range of PVDA scenarios. Empirical results demonstrate the state-of-the-art performance of our proposed PATAN across the multiple PVDA benchmarks.
    
\end{abstract}

\section{Introduction}
\label{section:intro}

Video-based problems have long been studied thanks to their wide applications in various fields. Neural networks have made notable advances in these problems with the availability of large-scale labeled video data. However, sufficiently large-scale training video data is sometimes unavailable, as annotations of video data are costly. 
Various \textit{Video-based Unsupervised Domain Adaptation} (VUDA) methods have been proposed to enable networks transfer knowledge from a labeled source domain to an unlabeled target domain by learning domain-invariant feature representations in the absence of target labels.

\begin{figure}[t]
\begin{center}
   \includegraphics[width=1.\linewidth]{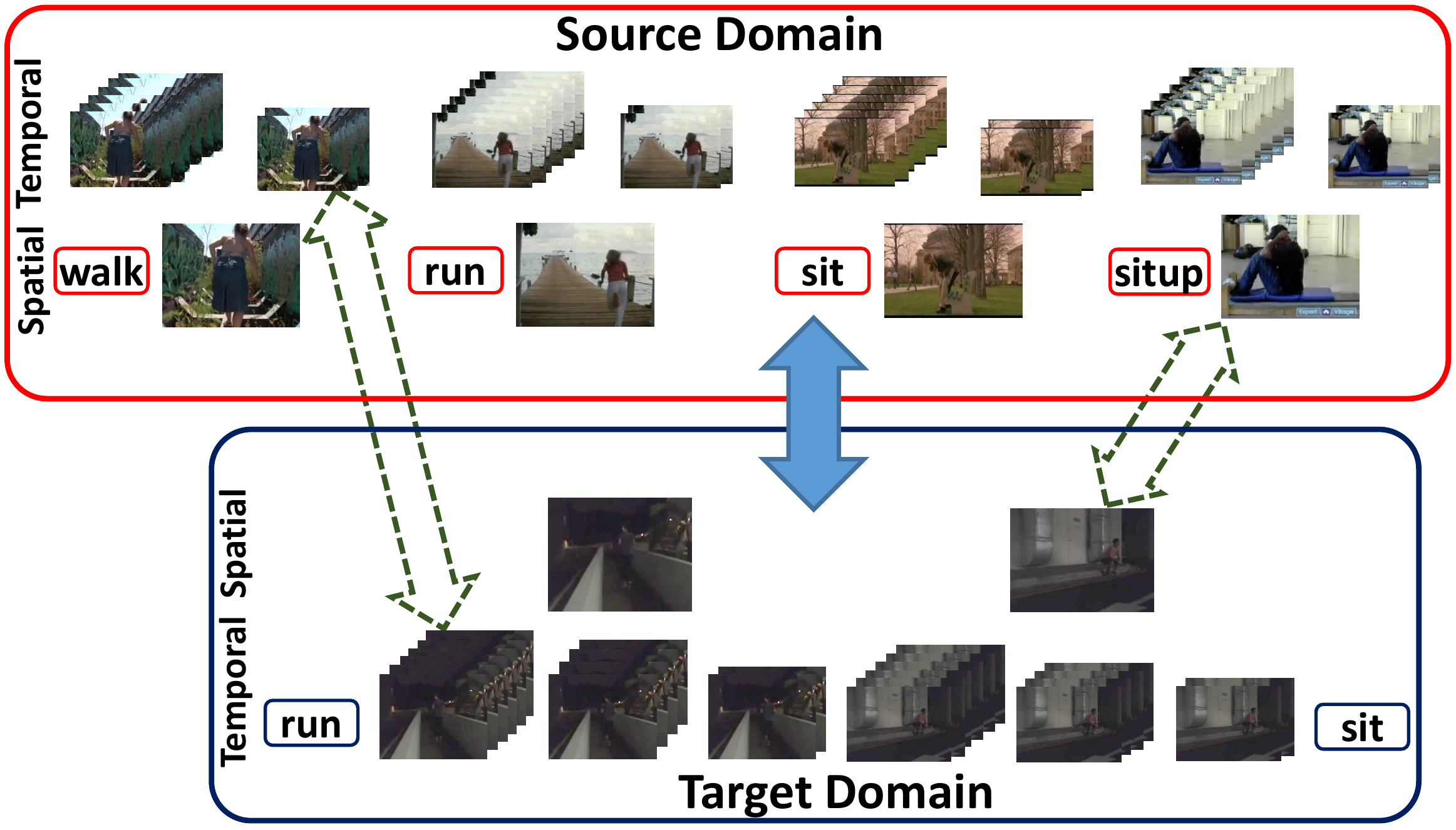}
\end{center}
   \caption{PVDA is a more general setting where the source label space subsumes the target label space. The key challenge of PVDA is the negative transfer caused by outlier source-only classes (`walk' and `situp'), with extra probability triggered by the incorrect alignment of target temporal features to the source temporal features of the outlier classes, depicted as the 
  left
   dashed arrow between videos from classes `run' and `walk'.}
\label{figure:1-1-intro}
\end{figure}

Though existing VUDA methods enable the learning of transferable features across domains, they generally assume that the video source and target domains share an identical label space, which may not hold in real-world applications. 
With the presence of large-scale labeled public video datasets,
it is more feasible to transfer representations learned in these datasets to unlabeled small-scale datasets.
Such a scenario is defined as \textit{Partial Domain Adaptation} (PDA), which relaxes the constraint of identical source and target label spaces by assuming that the target label space is a subspace of the source one. This assumption is more practical since large-scale public video datasets can subsume categories of the small-scale target datasets. However, the PDA problem is more challenging, since source-only classes may negatively influence the distribution alignment of target data, causing \textit{negative transfer}.

Compared to images that only contain spatial features, videos contain additional temporal features.
This leads to a novel \textit{Partial Video Domain Adaptation} (PVDA) problem, with trained networks transferred from video source domain to target domain, with the label space of video target domain being the subspace of the video source domain.
When transferring networks for PVDA, negative transfer would be triggered due to the possible spatial-temporal domain shift as depicted in Figure~\ref{figure:1-1-intro}, where the appearances of videos in class `walk' are different from that of videos in class `run', i.e.\ spatial features are different among videos in the two classes. However, videos from both classes share similar motion patterns where the actor moves further away from the camera in an upright position, indicating similar temporal features among the videos. When performing data distribution alignment, the similarities in temporal features would lead to videos in class `run' of the target domain to incorrectly align with videos in class `walk' of the source domain, triggering negative transfer.

A crucial step for tackling negative transfer in PVDA is the filtration of source-only outlier classes.
Different from images, temporal features should be leveraged for PVDA from two perspectives:
on one hand, effective temporal features should be constructed such that temporal features in outlier source-only classes discriminates those in target classes, alleviating the possibility of triggering negative transfer by temporal features; 
on the other hand, the temporal features should also contribute towards the filtration of source-only classes while eliminating possible mistakes caused by mis-classification of spatial features. 
To this end, we propose a \textbf{Partial Adversarial Temporal Attentive Network (PATAN)} to address the two challenges uniformly. 
\textbf{PATAN} first constructs robust overall temporal features by attentive combination of local temporal features which contain different aspects of the whole motion. 
The attentive combination builds upon the contribution of the local temporal feature towards the class filtration process where source-only classes are filtered. The constructed temporal feature would therefore have higher discriminability over source-only and target classes. Further, \textbf{PATAN} mitigates negative transfer in PDA through a class filtration process by utilizing local and overall temporal features jointly, alleviating possible mistakes during the class filtration process brought by the spatial features.


To further facilitate PVDA research, we propose three sets of benchmarks, built from widely used public datasets and a recent video dataset dedicated to low-illumination videos. The benchmarks proposed are: (a) \textit{UCF-HMDB\textsubscript{partial}}, (b) \textit{MiniKinetics-UCF}, and (c) \textit{HMDB-ARID\textsubscript{partial}}. The proposed datasets cover a wide range of PVDA scenarios, providing adequate baseline environment with distinct domain shift.

In summary, our contributions are three-fold. First, we formulated a novel and challenging \textit{Partial Video Domain Adaptation} (PVDA) problem. To the best of our knowledge, this is the first research that explores partial transfer in videos. 
Secondly, we analyze the challenges underlying PVDA and introduce PATAN to address the challenges. 
PATAN constructs robust temporal features,
while utilizing both spatial and temporal features for accurate class filtration.
Finally, we introduce several PVDA benchmarks, and demonstrate the effectiveness of our proposed method, achieving state-of-the-art performance across the multiple PVDA benchmarks proposed.

\section{Related Work}
\label{section:related}

\textbf{Unsupervised Domain Adaptation.} 
Unsupervised Domain Adaptation (UDA) aims to distill shared knowledge across labeled source domain and unlabeled target domain, improving the transferability of models. With the success of Generative Adversarial Network (GAN)~\cite{goodfellow2014generative}, researchers have proposed to align the cross-domain data distributions with additional domain discriminators that are trained with the feature generators in an adversarial manner~\cite{huang2011adversarial}, and construct adversarial loss~\cite{ganin2015unsupervised} for UDA. Subsequently, various adversarial-based UDA methods~\cite{tzeng2017adversarial,hoffman2018cycada,zou2019consensus} have been proposed for a wide range of image-based tasks, such as image recognition~\cite{ganin2015unsupervised,tzeng2015simultaneous,zhang2017joint}, object detection~\cite{chen2018domain,cai2019exploring,zhu2019adapting} and semantic segmentation~\cite{zou2018unsupervised,vu2019dada,chen2019crdoco}. More recently, with the wide applications of videos in various fields, there has been increasing research for Video-based Unsupervised Domain Adaptation (VUDA). The success of constructing domain-invariant features with adversarial-based methods extends to VUDA. This gives rise to the introduction of various adversarial-based VUDA approaches including VUDA approaches for action recognition~\cite{chen2019temporal,choi2020shuffle,pan2020adversarial} and action segmentation~\cite{chen2020action}.

\textbf{Partial Domain Adaptation.}
While the approaches for UDA and VUDA advances rapidly, these approaches assume that source and target domains share the same label space. A more general scenario that relaxes such assumption is introduced. Partial Domain Adaptation (PDA)~\cite{cao2018selective} enables models to transfer knowledge from many-class domains to few-class domains. Currently, there are multiple efforts towards the PDA problem. Among these, Selective Adversarial Network (SAN)~\cite{cao2018selective} adopts a multi-discriminator domain adversarial network with a weighting mechanism to select out source-only classes. Partial Adversarial Domain Adaptation (PADA)~\cite{cao2018partial} improves SAN by employing a single discriminator adversarial network and further applies the class weight to the source classifier. More recently, Example Transfer Network (ETN)~\cite{cao2019learning} is introduced to quantify the transferability of source data in a progressive weighting scheme through a discriminative domain discriminator. Generally, the PDA approaches above mitigate negative transfer by filtering out source-only outlier classes during the data alignment process.

Despite the notable advances achieved in PDA, the approaches are all built for image-based PDA problems, whereas Partial Video Domain Adaptation (PVDA) has not been tackled. PVDA is more challenging given that negative transfer could be triggered by temporal features, unique for videos. We propose to tackle PVDA with a novel network that constructs robust temporal features while utilizing spatial-temporal features for accurate class filtration.

\section{Proposed Method}
\label{section:method}

In the scenario of \textit{Partial Video Domain Adaptation} (PVDA), we are given a source domain $\mathcal{D}_S=\{(V_{iS},y_{iS})\}^{n_S}_{i=1}$ with $n_S$ labeled videos associated with $|\mathcal{C}_S|$ classes, and a target domain $\mathcal{D}_T=\{V_{iT}\}^{n_T}_{i=1}$ with $n_T$ unlabeled videos associated with $|\mathcal{C}_T|$ classes. The PVDA scenario is more general than VUDA by assuming that the source label space $\mathcal{C}_S$ is a superset of the target label space $\mathcal{C}_T$, i.e.\ $\mathcal{C}_T\subset \mathcal{C}_S$. The source and target domains of PVDA are characterized by two underlying probability distributions $p$ and $q$ respectively, where $p\neq q$. We also have $p_{\mathcal{C}_T}\neq q$, where $p_{\mathcal{C}_T}$ denotes the distribution of the source domain data in label space $\mathcal{C}_T$ of target domain.

To tackle the PVDA problem, we aim to construct a network capable of learning transferable features across source and target domains and minimizing the target classification risk. Compared to VUDA, PVDA poses more challenges to the network due to the existence of outlier label space in the source domain $\mathcal{C}_S\backslash \mathcal{C}_T$, which causes negative transfer effect to the network's performance. Meanwhile, during the training of the network, only unlabeled target domain data are accessible. Hence the part of which $\mathcal{C}_S$ shares with $\mathcal{C}_T$ is unknown. Therefore the key towards mitigating negative effects lies in the class filtration process which filters out the outlier source-only classes.

Current PDA approaches are built for image-based PDA problems, where the negative transfer could only be triggered by the alignment
of spatial features. 
Whereas for videos, negative transfer could be additionally triggered by the alignment 
of temporal features (e.g.\ the alignment of target videos in `run' to source videos in `walk' as depicted in Figure~\ref{figure:1-1-intro}). 
Thanks to the fact that current feature extractors would pay more attention across the spatial dimension, current PDA approaches may not be sensitive to negative transfer caused by the incorrect alignment of temporal features.
Therefore, we propose a novel Partial Adversarial Temporal Attentive Network (PATAN), to enable partial domain adaptation in an adversarial manner while mitigating negative transfer utilizing attentive temporal features. 
We begin by reviewing adversarial-based partial domain adaptation approaches, followed by a detailed illustration of PATAN.

\subsection{Adversarial-based Partial Domain Adaptation}
\label{section:method:advda}
Domain adaptation (DA) is achieved by matching the feature distributions of the source and target domains. One major line of approaches learns the domain-invariant features $\mathbf{f}$ in an adversarial manner where additional domain discriminators are trained with the feature generators in a min-max fashion. More specifically, the parameters $\theta_{f}$ of the feature extractor $G_{f}$ are learned by maximizing the losses of the domain discriminator $G_{d}$, while the parameters $\theta_{d}$ of the domain discriminator $G_{d}$ are trained by minimizing the losses of the domain discriminators $G_{d}$. Additionally, the loss of the source classifier $G_{y}$ is also minimized. The overall objective of adversarial-based DA networks can be formulated as in~\cite{ganin2016domain}:
\begin{equation}
\label{eqn:method:op-adv}
    \begin{aligned}
    \mathcal{L}_{0}
    &= \frac{1}{n_S}\sum\limits_{z_{i}\in \mathcal{D}_S}
    L_{y}(G_{y}(G_{f}(z_{i})), y_{i})\\
    &- \frac{\lambda}{n_A}\sum\limits_{z_{i}\in \mathcal{D}_A}
    L_{d}(G_{d}(G_{f}(z_{i})), d_{i}),
    \end{aligned}
\end{equation}
where $z_{i}$ is an input data point, $\mathcal{D}_A = \mathcal{D}_S\cup \mathcal{D}_T$ is the union of source and target domains with $n_A=|\mathcal{D}_A|$, $d_{i}$ is the domain label of input $z_{i}$, and $\lambda$ is the trade-off for the domain loss $L_{d}$ with respect to the source classification loss $L_{y}$. Both losses are implemented as cross-entropy losses. The min-max optimization process is achieved by the connecting a Gradient Reverse Layer (GRL) to $G_{d}$.

While the aforementioned adversarial-based networks can be applied to standard DA tasks, yielding reliable results, their performance deteriorates for PDA tasks due to negative transfer caused by outlier source-only classes within the label space of $\mathcal{C}_S\backslash \mathcal{C}_T$. Hence a class filtration process is applied to filter out these outlier classes.

The end result of this class filtration process are class weights $\gamma_{l}$ for each source domain label $l\in \mathcal{C}_S$, indicating the probability of each class of label space $\mathcal{C}_S$ overlapping with label space $\mathcal{C}_T$. To obtain the class weights $\gamma$, it is observed that the output of the source classifier $G_{y}$ for data $z_{i}$ well represents the probability distribution of $z_{i}$ over the source label space $\mathcal{C}_S$. The probability of the target data assigned to labels with space overlapped between $\mathcal{C}_S$ and $\mathcal{C}_T$ should be significantly larger than the probability of the target data assigned to outlier classes with label space $\mathcal{C}_S\backslash \mathcal{C}_T$. Therefore, the class weights $\gamma$ are generally obtained by the label predictions of the target data through the source classifier $G_{y}$, which indicates the probability of assigning the target data to each source class, and is formulated as:
\begin{equation}
\label{eqn:method:pada-w}
    \gamma = \frac{1}{n_T}\sum\limits_{z_{i}\in \mathcal{D}_T}G_{y}(G_{f}(z_{i})) = \frac{1}{n_T}\sum\limits_{i=1}^{n_T}\hat{y}_{i},
\end{equation}
where $\hat{y}_{i}$ is the label prediction of target input data $z_{i}$.

To down-weigh the contributions of the data in outlier classes for PDA tasks, the class weights $\gamma$ are applied to the adversarial-based networks, yielding the objective of PDA networks formulated as in~\cite{cao2018partial}:
\begin{equation}
\label{eqn:method:op-pada}
\resizebox{1.\linewidth}{!}{$
    \begin{aligned}
    \mathcal{L}_{p}
    &= \frac{1}{n_{S}}\sum\limits_{z_{i}\in \mathcal{D}_S}
    \gamma_{y_{i}} [L_{y}(G_{y}(G_{f}(z_{i})), y_{i}) + \lambda L_{d}(G_{d}(G_{f}(z_{i})), d_{i})]\\
    &- \frac{\lambda}{n_T}\sum\limits_{z_{i}\in \mathcal{D}_T}
    L_{d}(G_{d}(G_{f}(z_{i})), d_{i}),
    \end{aligned}
$}
\end{equation}
where $y_{i}$ is the ground truth of input $z_{i}$ in the source domain, while $\gamma_{y_{i}}$ is the corresponding class weight.

\subsection{Partial Adversarial Temporal Attentive Network}
\label{section:method:patan}
\begin{figure*}[t]
\begin{center}
   \includegraphics[width=.9\linewidth]{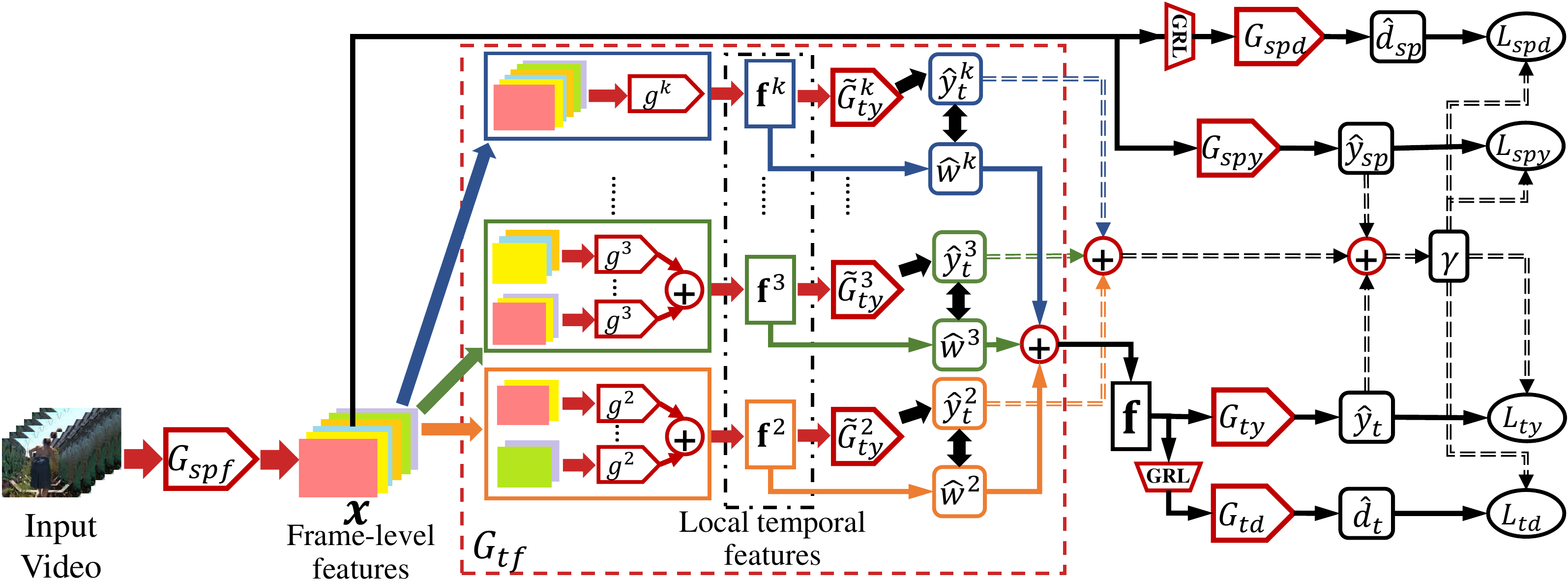}
\end{center}
    \caption{Architecture of the proposed PATAN. To mitigate negative transfer for PVDA effectively, robust overall feature $\mathbf{f}$ is constructed by weighted combination of local temporal features $\mathbf{f}^r$. The local temporal features $\mathbf{f}^r$ are built by fusing the time ordered frame-level features. The class weights of source domain classes $\gamma$ averages over the label predictions of the spatial feature, weighted local temporal features and the overall temporal feature of target data. $\gamma$ is applied to both the source domain label classifier and spatial/temporal domain discriminators. \textit{Best viewed in color and zoomed in.}}
\label{figure:3-2-patan}
\end{figure*}

Intuitively, when tackling the PVDA problem, the approach in Section~\ref{section:method:advda} could be directly integrated into videos, i.e.\ $z_{i}=V_{i}$. Given that videos contain both spatial and temporal features, one typical method for obtaining transferable video features is by separating the feature extractor $G_{f}$ into a spatial feature $G_{spf}$ and temporal feature extractor $G_{tf}$. The network constructed for PVDA could be formulated by simply substituting $G_{f}(z_{i})$ in Equation \ref{eqn:method:pada-w} and Equation \ref{eqn:method:op-pada} with $G_{tf}(G_{spf}(V_{i}))$.

One major drawback of direct integration of the above PDA approach into videos lies in the fact that video representations obtained through conventional video feature extractors (e.g.\ convolution-based networks) are mainly from the spatial features. The overall temporal information is generally encoded implicitly, usually implemented as a temporal pooling mechanism. 
Without explicit temporal features, the class filtration process in Section~\ref{section:method:advda} would depend mainly on the spatial features. Therefore the negative transfer may only be alleviated along the spatial dimension. 

In view of such drawback, we propose \textbf{Partial Adversarial Temporal Attentive Network (PATAN)} to mitigate negative transfer by utilizing spatial and temporal features jointly, as shown in Figure~\ref{figure:3-2-patan}. To utilize the temporal features of videos for negative transfer mitigation, the temporal features should be explicitly extracted first. Given the fact that humans can recognize actions by reasoning the observations across time, temporal features could be extracted utilizing Temporal Relation Module~\cite{zhou2018temporal}. We denote an input video with $k$ frames as $V_{i}=\{x_{i}^{(1)},x_{i}^{(2)},...,x_{i}^{(k)}\}$, where $x_{i}^{(j)}$ is the $j$th frame-level feature representation of the $i$th video obtained from the spatial feature extractor $G_{spf}$. The temporal feature of $V_{i}$ is constructed by a combination of multiple local temporal features, each built upon clips with $r$ temporal-ordered sampled frames where $r\in [2,k]$. Formally, a local temporal feature $\mathbf{f}_{i}^r$ is defined by:
\begin{equation}
\label{eqn:method:loc-t}
    \mathbf{f}_{i}^{r} = \sum\limits_{m} g^{r}((V_{i}^{r})_m),
\end{equation}
where $(V_{i}^{r})_m=\{x_{i}^{(a)},x_{i}^{(b)},...\}_m$ is the $m$th clip with $r$ temporal-ordered frames, with $a$ and $b$ denoting the frame indices. The local temporal feature of $\mathbf{f}_{i}^r$ is computed by fusing the time ordered frame-level features through function $g^{r}$, implemented as an Multi-Layer Perceptron (MLP).

Further, the key for mitigating negative transfer along the temporal dimension lies in the design of an effective class filtration process to down-weigh the effects of outlier classes with temporal features. The class filtration process is built upon the observation that the probability of the target data assigned to outlier classes with label space $\mathcal{C}_S\backslash \mathcal{C}_T$ should be significantly small. 
To make full use of the local temporal features
and to eliminate possible mis-assignment of target classes to spatial features
, we apply the above observation to all local temporal features of the target data. The label prediction of local temporal feature $\mathbf{f}_{i}^{r}$ is obtained as $\hat{y}_{ti}^{r} = \tilde{G}_{ty}^{r}(\mathbf{f}_{i}^{r})$, which gives the probability distribution of $\mathbf{f}_{i}^{r}$ across the source label space $\mathcal{C}_S$. Here $\tilde{G}_{ty}^{r}$ is the auxiliary source classifier for $\mathbf{f}_{i}^{r}$, and is trained as cross-entropy loss with source local temporal features, i.e.\ $\mathbf{f}_{i'}^r$ where $V_{i'}\in \mathcal{D}_S$.

To obtain the overall temporal feature and the class weights of each source class, one straight-forward strategy is to aggregate all local temporal features and their corresponding label predictions. However, not all local temporal features are equally important towards the mitigation of negative transfer. We introduce a \textit{label attention} mechanism to attend to local temporal features that contribute more toward the class filtration process. Specifically, the temporal features would be robust and the class filtration process would be effective only if temporal features in outlier source-only classes discriminates from those in target classes. 
If the features are of low discriminability and therefore the predictions are uncertain, the class weights of source classes which correlates with the predictions would be similar across all source classes. The network would be unable to filter out outlier source-only classes. Therefore, the proposed network should construct effective overall temporal features which attend to discriminable features that better distinguish if the label of the input data lies within the target label space $\mathcal{C}_T$ or the outlier label space $\mathcal{C}_S\backslash \mathcal{C}_T$. The certainty of the label prediction $\hat{y}_{ti}^{r}$ which corresponds to $\mathbf{f}_{i}^{r}$ is quantified by the additive inverse of the entropy of the label prediction as:
\begin{equation}
\label{eqn:method:loc-certainty}
    \mathbb{C}(\hat{y}_{ti}^{r})= \sum\limits_{c=1}^{|\mathcal{C}_S|} \hat{y}_{ti,c}^{r}log(\hat{y}_{ti,c}^{r}).
\end{equation}
For more stable optimization, a residual connection is added towards the formulation of the local temporal feature weight. The weight $\hat{w}_{i}^{r}$ of the local temporal feature $\mathbf{f}_{i}^{r}$ could therefore be generated as:
\begin{equation}
\label{eqn:method:loc-w}
    \hat{w}_{i}^{r} = \tanh(1 + \mathbb{C}(\hat{y}_{i}^{r})),
\end{equation}
where the $\tanh$ function is applied to ensure that weight $\hat{w}_{i}^{r}$ is constraint within a range of $[0,1]$. 

The weight $\hat{w}_{i}^{r}$ computed represents the contribution of the corresponding local temporal feature towards the class filtration process, which ultimately computes the class weights $\gamma_{l}$ for each source domain label $l\in \mathcal{C}_S$. The above \textit{label attention} weight is applied to both the generation of temporal attentive class weights utilizing the local temporal features and also the construction of the overall temporal feature. Formally, the overall temporal feature of input video $V_{i}$ with $k$ frames are constructed by:
\begin{equation}
\label{eqn:method:temp-feat}
    \mathbf{f}_{i} = G_{tf}(V_{i}) = \sum\limits_{r=2}^{k} \hat{w}_{i}^{r} \mathbf{f}_{i}^{r},
\end{equation}
where $G_{tf}$ denotes the overall temporal feature extractor as shown in Figure~\ref{figure:3-2-patan}. Meanwhile, the temporal attentive class weights generated for filtering out outlier source-only classes is formulated as:
\begin{equation}
\label{eqn:method:patan-w}
    \gamma = \frac{1}{n_T (k+1)}\sum\limits_{i=1}^{n_T}(\hat{y}_{ti} + \hat{y}_{spi}
    + \sum\limits_{r=2}^{k}\hat{w}_{i}^{r}\hat{y}_{i}^{r}),
\end{equation}
where the $\hat{y}_{ti}$ and $\hat{y}_{spi}$ are the label predictions of the $i$th input target video with temporal feature $\mathbf{f}_{i}$ and spatial feature $\mathbf{x}_{i}$, computed as $\hat{y}_{ti}=G_{ty}(\mathbf{f}_{i})$ and $\hat{y}_{spi}=G_{spy}(\mathbf{x}_{i})$. 

Finally, PATAN enables partial domain adaptation for videos by down-weighing the contributions of all source data belonging to the outlier label space $\mathcal{C}_S\backslash \mathcal{C}_T$. This is achieved by applying the temporal attentive class weight $\gamma$ to the source label classifier as well as the spatial and temporal domain discriminators over the source domain data. The overall optimization objective of the proposed PATAN is formulated as:
\begin{align}
\label{eqn:method:op-patan}
    \mathcal{L}
    &= \frac{1}{n_{S}}\sum\limits_{V_{i}\in \mathcal{D}_S}
    \gamma_{y_{i}} L_{ty}(G_{ty}(G_{tf}(G_{spf}(V_{i}))), y_{i}) \nonumber \\
    &+ \frac{1}{n_{S}}\sum\limits_{V_{i}\in \mathcal{D}_S}
    \gamma_{y_{i}} L_{spy}(G_{spy}(G_{spf}(V_{i})), y_{i}) \nonumber \\
    &- \frac{\lambda_{sp}}{n_S}\sum\limits_{V_{i}\in \mathcal{D}_S}
    \gamma_{y_{i}} L_{spd}(G_{spd}(G_{spf}(V_{i})), d_{i}) \nonumber \\
    &- \frac{\lambda_{sp}}{n_S}\sum\limits_{V_{i}\in \mathcal{D}_S}
    \gamma_{y_{i}} L_{td}(G_{td}(G_{tf}(G_{spf}(V_{i}))), d_{i}) \nonumber \\
    &- \frac{\lambda_t}{n_T}\sum\limits_{V_{i}\in \mathcal{D}_T}
    L_{spd}(G_{spd}(G_{spf}(V_{i})), d_{i}) \nonumber \\
    &- \frac{\lambda_t}{n_T}\sum\limits_{V_{i}\in \mathcal{D}_T}
    L_{td}(G_{td}(G_{tf}(G_{spf}(V_{i}))), d_{i}),
\end{align}
where $y_{i}$ is the ground truth of input $V_{i}$ in the source domain, while $\gamma_{y_{i}}$ is the corresponding class weight, and $\lambda_{sp}$ and $\lambda_{t}$ are the trade-offs for the domain loss $L_{spd}$ and $L_{td}$ with respect to the source classification losses $L_{ty}$ and $L_{spy}$.

\section{PVDA Benchmarks}
\label{section:pvda-db}

There are very limited cross-domain benchmark datasets for VUDA. Current cross-domain VUDA datasets are designed for the standard VUDA tasks, with the source label space constraint to be the same as target label space. To further facilitate PVDA research, we propose three sets of benchmarks, UCF-HMDB\textsubscript{\textit{partial}}, MiniKinetics-UCF, and HMDB-ARID\textsubscript{\textit{partial}}, which cover a wide range of PVDA scenarios and provide adequate baseline environment with distinct domain shift to facilitate PVDA research.

\textbf{UCF-HMDB\textsubscript{\textit{partial}}.}
UCF-HMDB\textsubscript{\textit{partial}} is constructed from two widely used video datasets: UCF101 (\textbf{U})~\cite{soomro2012ucf101} and HMDB51 (\textbf{H})~\cite{kuehne2011hmdb}. The overlapping classes between the two datasets are collected, resulting in 14 classes with 2,780 videos. The first 7 categories in alphabetic order of the target domain are chosen as target categories, and we construct two PVDA tasks: \textbf{U-14}$\to$\textbf{H-7} and \textbf{H-14}$\to$\textbf{U-7}. We follow the official split for the training and validation sets.

\textbf{MiniKinetics-UCF.}
MiniKinetics-UCF is built from two large-scale video datasets: MiniKinetics-200 (\textbf{M})~\cite{xie2017rethinking} and UCF101 (\textbf{U})~\cite{soomro2012ucf101}, which contains 45 overlapping classes. Similar to the construction of UCF-HMDB\textsubscript{\textit{partial}}, the first 18 categories in alphabetic order of the target domain are chosen as target categories, resulting in two PVDA tasks: \textbf{M-45}$\to$\textbf{U-18} and \textbf{U-45}$\to$\textbf{M-18}. In this dataset, there are a total of 22,102 videos, nearly 8 times larger than that of UCF-HMDB\textsubscript{\textit{partial}}. Thus this dataset could validate the effectiveness of PVDA approaches on large-scale dataset.

\textbf{HMDB-ARID\textsubscript{\textit{partial}}.}
HMDB-ARID\textsubscript{\textit{partial}} is constructed with the goal of leveraging current video datasets to boost performance on videos shot in adverse environments. It incorporates both HMDB51 (\textbf{H})~\cite{kuehne2011hmdb} and a more recent dark dataset, ARID (\textbf{A})~\cite{xu2020arid}, with videos shot under adverse illumination conditions. Statistically, videos in ARID possess much lower RGB mean value and standard deviation (std), which leads larger domain shift between ARID and HMDB51 compared to other cross-domain datasets. The overlapping classes between the two datasets are collected, resulting in 10 classes with 3,252 videos. The first 5 categories in alphabetic order of the target domain are chosen as target categories, resulting in two PVDA tasks: \textbf{H-10}$\to$\textbf{A-5} and \textbf{A-10}$\to$\textbf{H-5}. For all the aforementioned benchmarks, the training and validation sets are separated following the official split methods.

\begin{table*}[!ht]
\center
\resizebox{.9\linewidth}{!}{\noindent
\begin{tabular}{c|c|cc|cc|cc}
\hline
\hline
  \multicolumn{2}{c|}{\multirow{2}{*}{Methods}} &
  \multicolumn{2}{c|}{UCF-HMDB\textsubscript{\textit{partial}}} &
  \multicolumn{2}{c|}{MiniKinetics-UCF} &
  \multicolumn{2}{c}{HMDB-ARID\textsubscript{\textit{partial}}}\\
\cline{3-8}
\multicolumn{2}{c|}{} & \textbf{U-14}$\to$\textbf{H-7} & \textbf{H-14}$\to$\textbf{U-7} & \textbf{M-45}$\to$\textbf{U-18} & \textbf{U-45}$\to$\textbf{M-18} & \textbf{H-10}$\to$\textbf{A-5} & \textbf{A-10}$\to$\textbf{H-5} \\
\hline
\parbox{0.12\linewidth}{\centering Source-only}
& TRN~\cite{zhou2018temporal}
& 62.85\% & 78.95\% & 78.77\% & 54.14\% & 14.10\% & 26.00\% \\
\hline
\multirow{4}{*}{\parbox{0.12\linewidth}{\centering Adversarial-based}}
& DANN~\cite{ganin2015unsupervised}
& 60.95\% & 74.44\% & 79.21\% & 52.25\% & 20.77\% & 12.00\% \\
& TA\textsuperscript{3}N~\cite{chen2019temporal}
& 50.49\% & 70.68\% & 75.70\% & 48.23\% & 18.30\% & 24.00\% \\
& PADA~\cite{cao2018partial}
& 65.71\% & 82.33\% & 82.43\% & 61.23\% & 21.79\% & 30.67\% \\
& ETN~\cite{cao2019learning}
& 67.88\% & 82.89\% & 83.33\% & 62.51\% & 21.40\% & 28.82\% \\
\hline
\multirow{3}{*}{\parbox{0.12\linewidth}{\centering Discrepancy-based}}
& MK-MMD~\cite{long2015learning}
& 58.57\% & 82.71\% & 79.79\% & 55.79\% & 21.28\% & 14.00\% \\
& MCD~\cite{saito2018maximum}
& 55.71\% & 73.31\% & 75.13\% & 52.48\% & 12.56\% & 14.67\% \\
& MDD~\cite{zhang2019bridging}
& 62.58\% & 80.45\% & 80.12\% & 50.35\% & 15.13\% & 9.33\% \\
\hline
Ours
& \textbf{PATAN}     & \textbf{73.81\%} & \textbf{89.85\%} & \textbf{86.82\%} & \textbf{65.25\%} & \textbf{26.41\%} & \textbf{34.67\%} \\
\hline
\hline
\end{tabular}
}
\smallskip
\caption{Results for Partial Video Domain Adaptation on UCF-HMDB\textsubscript{\textit{partial}}, MiniKinetics-UCF and HMDB-ARID\textsubscript{\textit{partial}}.}
\label{table:5-1-sota}
\end{table*}
\section{Experiments}
\label{section:exps}

\begin{figure*}[t]
\begin{center}
   \includegraphics[width=.85\linewidth]{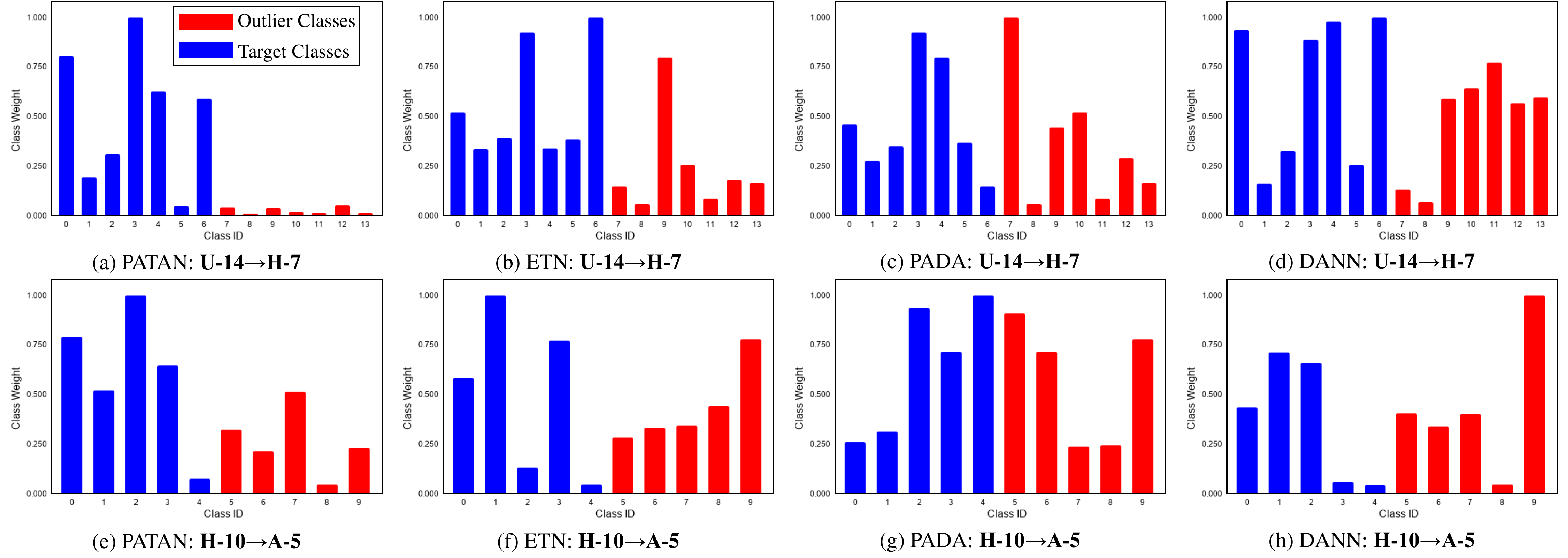}
\end{center}
    \caption{Histograms of class weights learned by PATAN, ETN, PADA and DANN on settings \textbf{U-14}$\to$\textbf{H-7} and \textbf{H-10}$\to$\textbf{A-5}.}
\label{figure:5-3-gamma}
\end{figure*}

\begin{table}[t]
\center
\resizebox{.85\linewidth}{!}{
\smallskip\begin{tabular}{c|cc}
\hline
\hline
Methods & \textbf{U-14}$\to$\textbf{H-7} & \textbf{H-14}$\to$\textbf{U-7} \\
\hline
\textbf{PATAN} & \textbf{73.81\%} & \textbf{89.85\%}\\
\hline
PATAN w/o attentive & 71.43\% & 85.34\%\\
PATAN w/o local weights & 70.47\% & 84.21\%\\
PATAN w/o classifier & 69.52\% & 82.71\%\\
PATAN w/o adversarial & 67.14\% & 81.58\%\\
\hline
\hline
\end{tabular}
}
\smallskip
\caption{Ablation studies of PATAN on UCF-HMDB\textsubscript{\textit{partial}}.}
\label{table:5-2-efficacy}
\end{table}

In this section, we evaluate our proposed PATAN by performing cross-domain action recognition on PVDA benchmarks introduced in Section~\ref{section:pvda-db}. We present state-of-the-art results on all proposed benchmarks. We also present ablation studies and empirical analysis of our proposed network to verify our design. \textit{Code is provided in the appendix.}

\subsection{Experimental Settings}
\label{section:exps:settings}
We perform action recognition tasks on all three benchmarks: UCF-HMDB\textsubscript{\textit{partial}}, MiniKinetics-UCF and HMDB-ARID\textsubscript{\textit{partial}}, with a total of six cross-domain settings as suggested in Section~\ref{section:pvda-db}. For all six settings, we use all labeled source videos and all unlabeled target videos for PVDA following standard evaluation protocols~\cite{saenko2010adapting,long2015learning}. We report the top-1 accuracy on the target datasets. Our experiments are implemented using PyTorch~\cite{paszke2019pytorch} library. All methods utilize the Temporal Relation Network (TRN)~\cite{zhou2018temporal} as the backbone for video feature extraction, with the model pretrained on ImageNet~\cite{deng2009imagenet}. A more detailed implementation specification could be found in the appendix.

\subsection{Overall Results and Comparisons}
\label{section:exps:results}
We compare the performance of PATAN with competitive and state-of-the-art UDA/VUDA approaches and state-of-the-art PDA approaches. These include: (a) adversarial-based methods: DANN~\cite{ganin2015unsupervised}, 
TA\textsuperscript{3}N~\cite{chen2019temporal},
PADA~\cite{cao2018partial} and ETN~\cite{cao2019learning}; and (b) discrepancy-based methods: MK-MMD~\cite{long2015learning}, MCD~\cite{saito2018maximum} and MDD~\cite{zhang2019bridging}. Additionally, we report the results of the backbone feature extractor TRN, where TRN is trained with supervised source data only and validated on the target data. Table~\ref{table:5-1-sota} shows the comparison of performances between our proposed PATAN and the methods as mentioned in all six PVDA settings.

The results in Table~\ref{table:5-1-sota} show that our proposed PATAN achieves the best results on all six settings, and substantially outperforms previous approaches by noticeable margins. It can be observed that for all UDA/VUDA approaches, i.e.\ DANN, TA\textsuperscript{3}N, MK-MMD, MCD, and MDD, there exists at least three settings where their performances are inferior to that of TRN trained without any domain adaptation methods. This suggests that these methods suffer from the negative transfer issue of PVDA. 

Compare to previous PDA approaches PADA and ETN, our proposed PATAN exceeds both approaches consistently, with an average $11.27\%$ relative improvement towards PADA, and an average $11.57\%$ relative improvement towards ETN. These large improvements imply the effectiveness of building temporal attentive features and incorporating local and overall temporal features for class filtration. In particular, the improvement of PATAN with respect to PADA and ETN is most significant for HMDB-ARID\textsubscript{\textit{partial}}, with a relatively average improvement of $17.12\%$ and $21.85\%$ towards PADA and ETN respectively. HMDB-ARID\textsubscript{\textit{partial}} possesses the largest domain shift across the source and target domains, with the lowest source-only accuracies. This suggests that the class filtration process may not be accurate by utilizing any single feature, which explains the relatively small improvement in performances of PADA and ETN compared to UDA/VUDA approaches. On the contrary, by constructing temporal features by \textit{label attention}, the effectiveness of the class filtration process is improved with features of higher certainty attended, thus explains the large improvements brought by PATAN.

\subsection{Ablation Studies}
\label{section:exps:ablation}
To go deeper with the efficacy of the proposed PATAN network, we perform ablation studies by evaluating PATAN against its variants: (a) \textbf{PATAN w/o attentive} is the variant where the \textit{label attention} weight is not computed, therefore the overall temporal feature and class weights of source classes are the result of the aggregation of all local temporal features and the label predictions of all local and overall temporal features with that of the spatial feature; (b) \textbf{PATAN w/o local weights} is the variant where the class weights $\gamma$ do not incorporate all the local weights; (c) \textbf{PATAN w/o classifier} is the variant without the class weights $\gamma$ applied on the source spatial and temporal classifiers, and (d) \textbf{PATAN w/o adversarial} is the variant without the class weights applied on the spatial and temporal domain discriminators. The results of the variants are presented in Table~\ref{table:5-2-efficacy}.

Specifically, PATAN outperforms PATAN w/o attentive by a noticeable margin proves the necessity of combining local temporal features and class weights with \textit{label attention}, which constructs more discriminable overall temporal features. Similarly, PATAN's superior performance over PATAN w/o local weights proves that the class filtration process could be improved by utilizing label prediction of local temporal features. We note that the results of both PATAN w/o attentive and PATAN w/o local weights outperform that of PADA and ETN. This further justifies the effectiveness of both utilizing local temporal features for class filtration and constructing attentive overall temporal features with \textit{label attention}. Further, PATAN outperforms both PATAN w/o classifier and PATAN w/o adversarial by huge margins. This strongly suggests that the class weights applied can assign small weights on outlier classes and down-weigh the source data of the outlier classes effectively, the class weights applied thus mitigates negative transfer and boost the performance for PVDA.

\begin{figure*}[t]
\begin{center}
   \includegraphics[width=.95\linewidth]{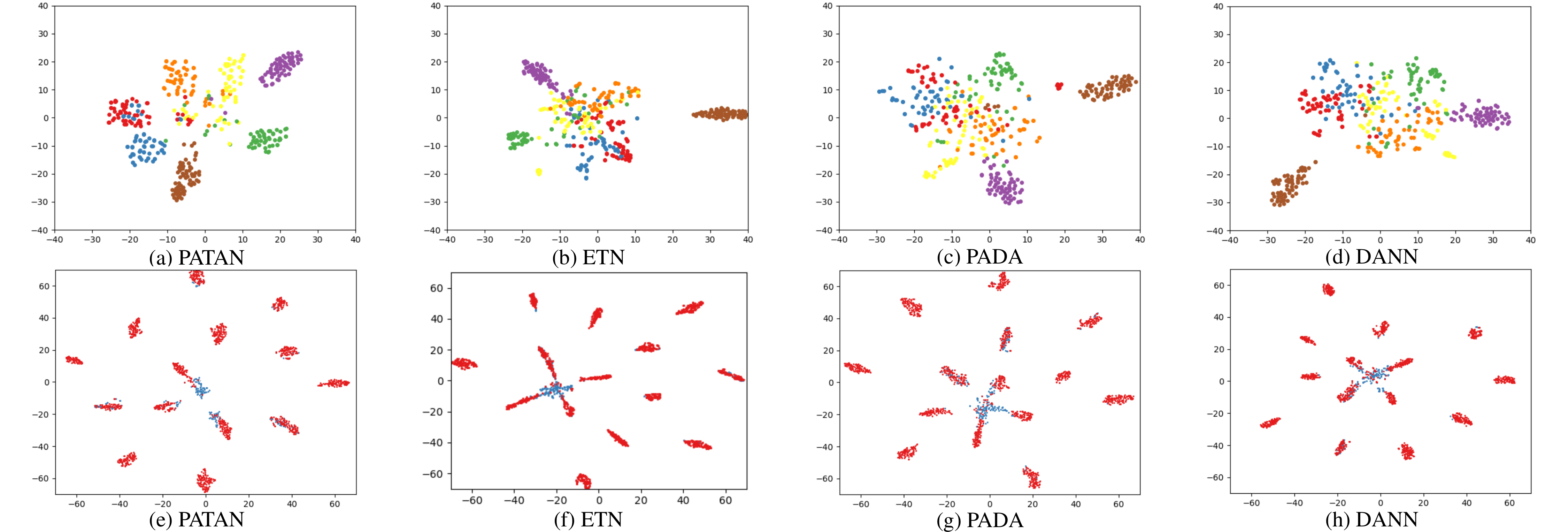}
\end{center}
    \caption{Visualization of features learned by PATAN, ETN, PADA, and DANN, with class information ((a)-(d)) and domain information ((e)-(h)). Different classes are denoted by different colors. The red dots represent data from the source domain while the blue dots represent data from the target domain.}
\label{figure:5-4-tsne}
\end{figure*}

\begin{figure}[t]
\begin{center}
   \includegraphics[width=.8\linewidth]{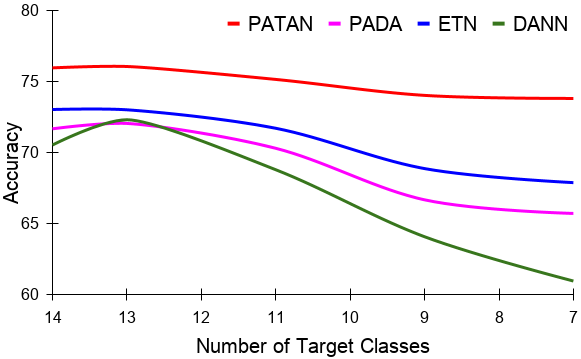}
\end{center}
   \caption{Accuracy with different number of target classes.}
\label{figure:5-5-target}
\end{figure}

\subsection{Empirical Analysis}
To further understand our proposed PATAN, we perform empirical analysis focusing on four areas of interest: class weights visualization, affect of number of target classes and feature visualization.

\textbf{Class weights visualization.}
We first illustrate and compare the learned class weights $\gamma$ generated by methods PATAN, ETN, PADA and DANN for settings \textbf{U-14}$\to$\textbf{H-7} and \textbf{H-10}$\to$\textbf{A-5} in Figure~\ref{figure:5-3-gamma}. It could be observed that our proposed PATAN assigns much smaller weights to the outlier source only classes than to the shared target classes, which shows that PATAN could effectively filter out the outlier classes. It is noted that the difference of class weights between target and outlier classes is less significant for \textbf{H-10}$\to$\textbf{A-5}, given the much larger cross-domain shift for dataset HMDB-ARID\textsubscript{\textit{partial}}. Despite the difficulty brought by the large domain shift, our proposed PATAN still assigns significantly larger weights for target classes compared to other methods. The much larger weights assigned to target classes show that our network could effectively filter out outlier classes, therefore explains the strong performance of PATAN on both datasets. Though both PADA and ETN incorporate class filtration processes to filter outlier classes, the effectiveness of such a process is hindered by failing to incorporate temporal features. This results in much poorer performances.

\textbf{Affect of number of target classes.}
We investigate a wider spectrum of PVDA by varying the number of target classes, conducted with the UCF-HMDB\textsubscript{\textit{partial}} dataset. The result of the accuracy of the target dataset against the different numbers of target classes is shown in Figure~\ref{figure:5-5-target}. it is observed that the performance of DANN, PADA, and ETN degrades noticeably with fewer target classes. This is a clear indication of negative transfer brought by the increasing outlier classes. Comparatively, the performance of PATAN is more stable and is consistently better than all compared methods. The stability of performance suggests that PATAN effectively alleviates the influence of outlier classes. It could also be observed that when the number of target classes is equivalent to that of source classes (in this case 14), the PVDA task is turned into a standard VUDA task. Under this condition, our PATAN also performs better than DANN. This shows that the class filtration process will not degrade performance when there are no outlier classes. 


\textbf{Feature visualization.}
We further plot the t-SNE embeddings~\cite{van2008visualizing} of the features learned by PATAN, ETN, PADA, and DANN for the \textbf{U-14}$\to$\textbf{H-7} with class information in the target domain as shown in Figure~\ref{figure:5-4-tsne} (a)-(d), and with domain information as shown in Figure~\ref{figure:5-4-tsne} (e)-(h). From Figure~\ref{figure:5-4-tsne} (a), it is observed that the features learned by PATAN are more clustered. This proves that features extracted by PATAN with \textit{label attention} have higher discriminability. Meanwhile, Figure~\ref{figure:5-4-tsne} (f)-(h) shows that other methods align target data to all source classes, which includes outlier ones, triggering negative transfer. We note that though ETN and PADA include class filtration processes, the negative transfer is still triggered due to misalignment of temporal features not utilized in their class filtration processes. Comparatively, PATAN only aligns target data to the shared classes (7 classes), alleviating the effects of the outlier classes.

\section{Conclusion}
\label{section:concl}

In this work, we propose a novel approach for partial video domain adaptation (PVDA). Unlike previous approaches where only spatial features are utilized for mitigating negative transfer in partial domain adaptation, the new PATAN tackles PVDA with full utilization of both spatial and temporal features, filtering out outlier source-only classes effectively. The proposed PATAN also attends to local temporal features that contribute more towards the class filtration process. We further introduce novel PVDA benchmarks to facilitate PVDA research, which are the first PVDA benchmarks introduced. Our proposed PATAN addresses the PVDA problem well, justified by extensive experiments across the proposed PVDA benchmarks.

\section{Supplementary: PVDA Benchmarks}
\label{section:supp:pvda-db}

In this work, we propose three sets of benchmarks, UCF-HMDB\textsubscript{\textit{partial}}, MiniKinetics-UCF, and HMDB-ARID\textsubscript{\textit{partial}}, which cover a wide range of \textit{Partial Video Domain Adaptation} (PVDA) scenarios and provide adequate baseline environment with distinct domain shift to facilitate PVDA research. Here we provide more detail on each benchmark.

\begin{figure}[t]
\begin{center}
   \includegraphics[width=1.\linewidth]{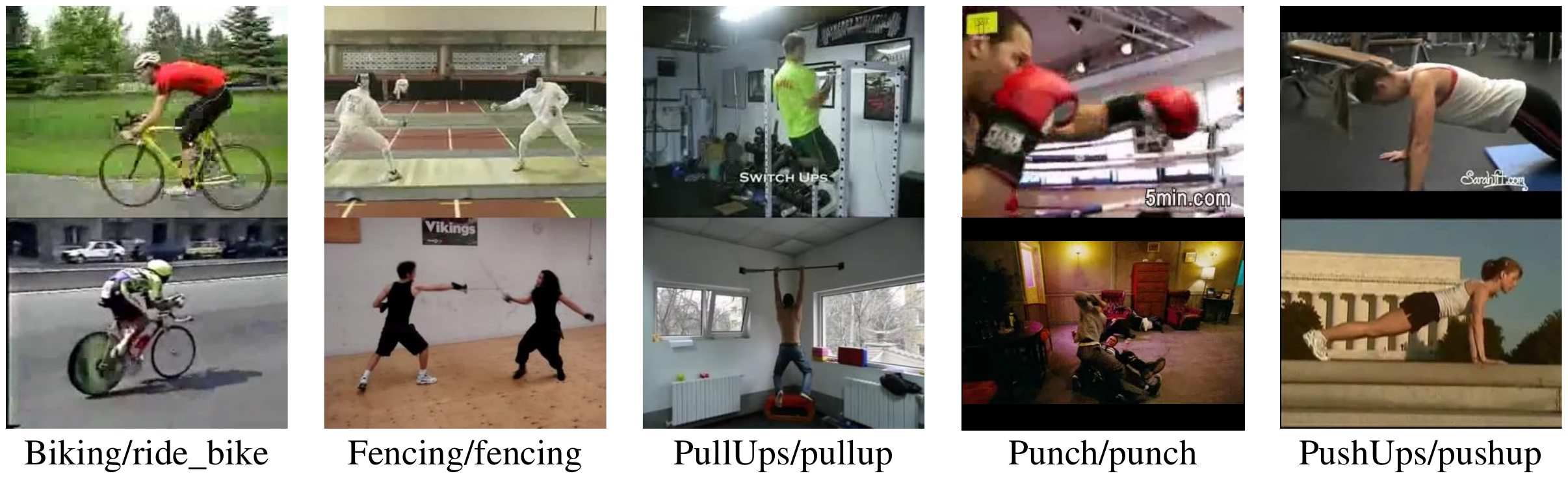}
\end{center}
   \caption{Sampled frames of videos from classes in UCF-HMDB\textsubscript{\textit{partial}}. Sampled frames from UCF101 are shown in the upper row, and those from HMDB51 are shown in the lower row.}
\label{figure:s1-1-uh}
\end{figure}

\paragraph{UCF-HMDB\textsubscript{\textit{partial}}.}
UCF-HMDB\textsubscript{\textit{partial}} is built from two widely used video datasets: UCF101 (\textbf{U})~\cite{soomro2012ucf101} and HMDB51 (\textbf{H})~\cite{kuehne2011hmdb}. The overlapping classes between the two datasets are collected, resulting in 14 classes with 2,780 videos. Among which are 980 training videos and 210 testing videos from HMDB51, 1,324 training videos and 266 testing videos from UCF101. The list of the 14 overlapping classes are listed in Table~\ref{table:s-1-uhpartial}. The first 7 categories in alphabetic order of the target domain are chosen as target categories, and we construct two PVDA tasks: \textbf{U-14}$\to$\textbf{H-7} and \textbf{H-14}$\to$\textbf{U-7}. We follow the official split for the training and validation sets. Figure~\ref{figure:s1-1-uh} shows the comparison of sampled frames from UCF-HMDB\textsubscript{\textit{partial}}.

\begin{table}[t]
\center
\smallskip\begin{tabular}{c|c}
\hline
\hline
UCF101 Class & HMDB51 Class \\
\hline
RockClimbingIndoor & climb \\
Diving & dive \\
Fencing & fencing \\
GolfSwing & golf \\
HandstandWalking & handstand \\
SoccerPenalty & kick\textunderscore ball \\
PullUps & pullup \\
Punch & punch \\
PushUps & pushup \\
Biking & ride\textunderscore bike \\
HorseRiding & ride\textunderscore horse \\
Basketball & shoot\textunderscore ball \\
Archery & shoot\textunderscore bow \\
WalkingWithDog & walk \\
\hline
\hline
\end{tabular}
\smallskip
\caption{List of overlapping classes between UCF101 and HMDB51.}
\label{table:s-1-uhpartial}
\end{table}

\begin{figure}[t]
\begin{center}
   \includegraphics[width=1.\linewidth]{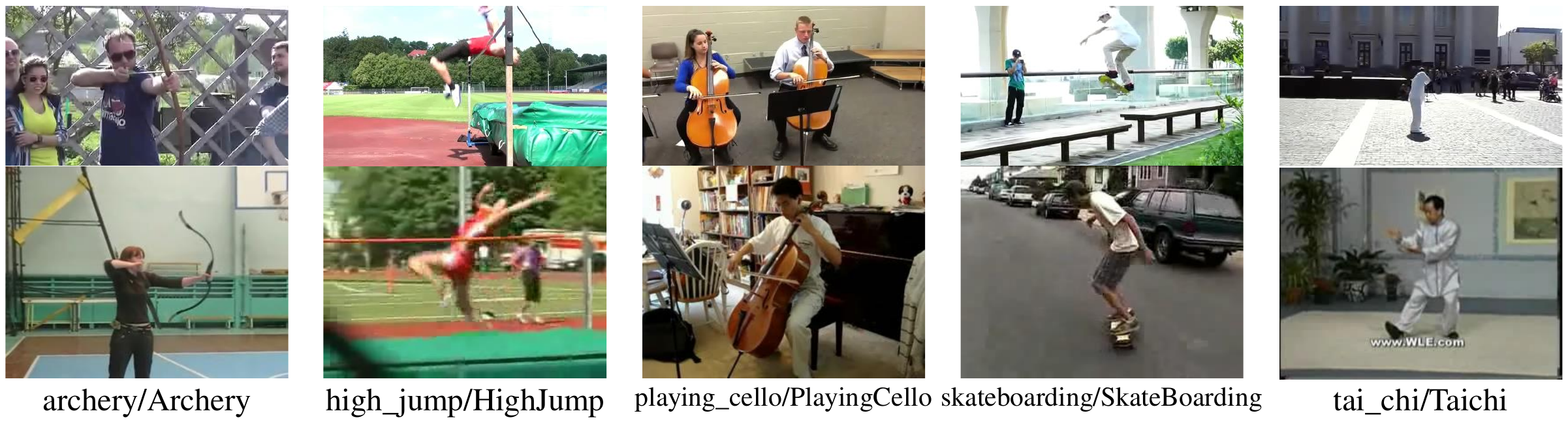}
\end{center}
   \caption{Sampled frames of videos from classes in MiniKinetics-UCF. Sampled frames from MiniKinetics-200 are shown in the upper row, while those from UCF101 are shown in the lower row.}
\label{figure:s1-2-mu}
\end{figure}

\paragraph{MiniKinetics-UCF.}
MiniKinetics-UCF is built from two large-scale video datasets: MiniKinetics-200 (\textbf{M})~\cite{xie2017rethinking} and UCF101 (\textbf{U})~\cite{soomro2012ucf101}. MiniKinetics-200 is a subset of the Kinetics \cite{kay2017kinetics} dataset, with 200 of its categories.
There are 45 overlapping classes between MiniKinetics-200 and UCF101, as shown in Table~\ref{table:s-2-mu}. Similar to the construction of UCF-HMDB\textsubscript{\textit{partial}}, the first 18 categories in alphabetic order of the target domain are chosen as target categories, resulting in two PVDA tasks: \textbf{M-45}$\to$\textbf{U-18} and \textbf{U-45}$\to$\textbf{M-18}. In this dataset, there are a total of 22,102 videos, with 4,253 training videos and 683 testing videos from UCF101, along with 16,743 training videos and 423 testing videos from MiniKinetics-200. The number of videos is nearly 8 times larger than that of UCF-HMDB\textsubscript{\textit{partial}}. Thus this dataset could validate the effectiveness of PVDA approaches on large-scale datasets. Figure~\ref{figure:s1-2-mu} shows the comparison of sampled frames from MiniKinetics-UCF.

\begin{table*}[t]
\center
\resizebox{1.\linewidth}{!}{
\smallskip\begin{tabular}{cc|cc|cc}
\hline
\hline
MiniKinetics-200 Class & UCF101 Class & MiniKinetics-200 Class & UCF101 Class & MiniKinetics-200 Class & UCF101 Class \\
\hline
archery & Archery & high\textunderscore jump & HighJump & pole\textunderscore vault & PoleVault \\
bench\textunderscore pressing & BenchPress & hula\textunderscore hooping & HulaHoop & pull\textunderscore ups & PullUps \\
biking\textunderscore through\textunderscore snow & Biking & javelin\textunderscore throw & JavelinThrow & riding\textunderscore or\textunderscore walking\textunderscore with\textunderscore horse & HorseRiding \\
blowing\textunderscore out\textunderscore candles & BlowingCandles & jetskiing & Skijet & rock\textunderscore climbing & RockClimbingIndoor \\
bowling & Bowling & juggling\textunderscore balls & JugglingBalls & salsa\textunderscore dancing & SalsaSpin \\
brushing\textunderscore teeth & BrushingTeeth& long\textunderscore jump & LongJump & shaving\textunderscore head & ShavingBeard \\
canoeing\textunderscore or\textunderscore kayaking & Kayaking & lunge & Lunges & shot\textunderscore put & Shotput \\
catching\textunderscore or\textunderscore throwing\textunderscore baseball & BaseballPitch & making\textunderscore pizza & PizzaTossing & skateboarding & SkateBoarding \\
catching\textunderscore or\textunderscore throwing\textunderscore frisbee & FrisbeeCatch & marching & BandMarching & skiing & Skiing \\
clean\textunderscore and\textunderscore jerk & CleanAndJerk& playing\textunderscore basketball & Basketball & squat & BodyWeightSquats \\
crawling\textunderscore baby & BabyCrawling & playing\textunderscore cello & PlayingCello & surfing\textunderscore water & Surfing \\
diving\textunderscore cliff & CliffDiving & playing\textunderscore guitar & PlayingGuitar & swimming\textunderscore breast\textunderscore stroke & BreastStroke \\
dunking\textunderscore basketball & BasketballDunk & playing\textunderscore tennis & TennisSwing & tai\textunderscore chi & Taichi \\
golf\textunderscore driving & GolfSwing & playing\textunderscore violin & PlayingViolin & throwing\textunderscore discus & ThrowDiscus \\
hammer\textunderscore throw & HammerThrow & playing\textunderscore volleyball & VolleyballSpiking & walking\textunderscore the\textunderscore dog & WalkingWithDog \\
\hline
\hline
\end{tabular}
}
\smallskip
\caption{List of overlapping classes between MiniKinetics-200 and UCF101.}
\label{table:s-2-mu}
\end{table*}

\begin{figure}[t]
\begin{center}
   \includegraphics[width=1.\linewidth]{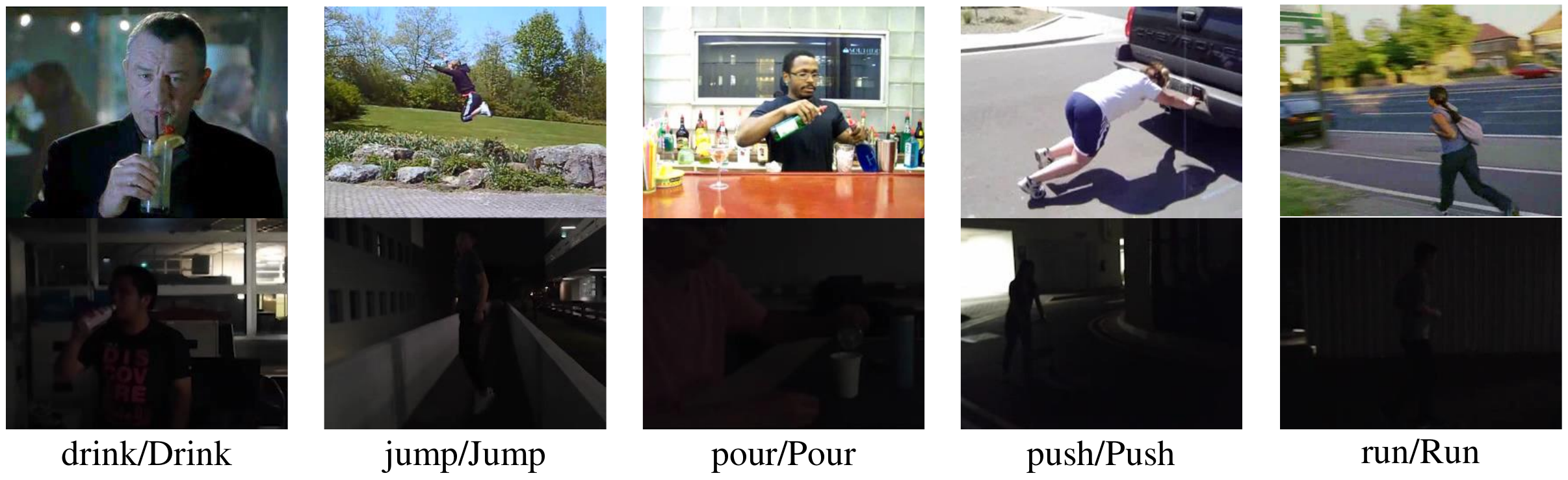}
\end{center}
   \caption{Sampled frames of videos from classes in HMDB-ARID\textsubscript{\textit{partial}}. Sampled frames from HMDB51 are shown in the upper row, while those from ARID are shown in the lower row.}
\label{figure:s1-3-ha}
\end{figure}

\paragraph{HMDB-ARID\textsubscript{\textit{partial}}.}
HMDB-ARID\textsubscript{\textit{partial}} is built with the goal of leveraging current video datasets to boost performance on videos shot in adverse environments. It incorporates both HMDB51 (\textbf{H})~\cite{kuehne2011hmdb} and a more recent dark dataset, ARID (\textbf{A})~\cite{xu2020arid}, with videos shot under adverse illumination conditions. Compared with current action recognition datasets (e.g.\ UCF101, HMDB51, MiniKinetics-200), videos in ARID are characterized by low brightness and low contrast. Statistically, videos in ARID possess much lower RGB mean value and standard deviation (std) as presented in Table~\ref{table:s-3-stat_compare}. This leads to larger domain shift between ARID and HMDB51 compared to other cross-domain datasets. 
The overlapping classes between the two datasets are collected, resulting in 10 classes with 3,252 videos, which includes 2,012 training videos and 390 testing videos from ARID, and 700 training videos and 150 testing videos from HMDB51. The list of the 10 overlapping classes is listed in Table~\ref{table:s-3-hapartial}. Similar to the other two PVDA benchmarks, the first 5 categories in alphabetic order of the target domain are chosen as target categories, resulting in two PVDA tasks: \textbf{H-10}$\to$\textbf{A-5} and \textbf{A-10}$\to$\textbf{H-5}. For all the aforementioned benchmarks, the training and validation sets are separated following the official split methods. Figure~\ref{figure:s1-3-ha} shows the comparison of sampled frames from HMDB-ARID\textsubscript{\textit{partial}}.

\begin{table}[t]
\center
\smallskip\begin{tabular}{c|c}
\hline
\hline
HMDB51 Class & ARID Class \\
\hline
RockClimbingIndoor & climb \\
Diving & dive \\
Fencing & fencing \\
GolfSwing & golf \\
HandstandWalking & handstand \\
SoccerPenalty & kick\textunderscore ball \\
PullUps & pullup \\
Punch & punch \\
PushUps & pushup \\
Biking & ride\textunderscore bike \\
\hline
\hline
\end{tabular}
\smallskip
\caption{List of overlapping classes between HMDB51 and ARID.}
\label{table:s-3-hapartial}
\end{table}

\begin{table}[t]
\centering
\resizebox{1.\linewidth}{!}{
\smallskip
\begin{tabular}{l|c|c}
\hline
\hline
Dataset & RGB Mean & RGB Std\\
\hline
HMDB51 & [0.424,0.364,0.319] & [0.268,0.255,0.260]\\
UCF101 & [0.409,0.397,0.358] & [0.266,0.265,0.270]\\
MiniKinetics-200 & [0.435,0.394,0.381] & [0.225,0.225,0.214]\\
\hline
ARID & [0.079,0.074,0.073] & [0.101,0.098,0.090]\\
\hline
\hline
\end{tabular}
}
\smallskip
\caption{Comparison of RGB mean and standard deviation (std) over common action recognition datasets and the ARID dataset.}
\label{table:s-3-stat_compare}
\end{table}

\section{Supplementary: Detailed Implementation of the Proposed Network}
\label{section:supp:detail-imp}

\begin{figure*}[t]
\begin{center}
   \includegraphics[width=.9\linewidth]{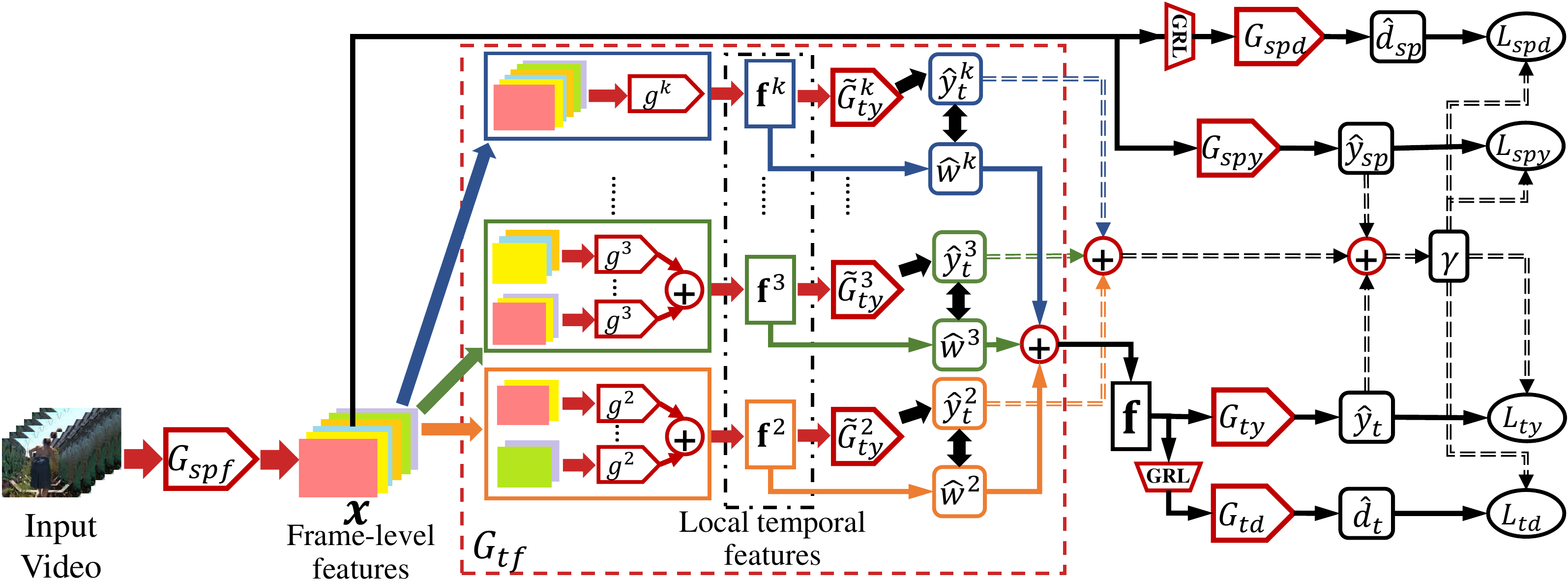}
\end{center}
    \smallskip\caption{Architecture of the proposed PATAN. \textit{Best viewed in color and zoomed in.}}
\label{figure:s2-3-patan}
\end{figure*}

As presented in Section 3, we propose PATAN to tackle the PVDA problem by constructing robust temporal features and utilizing both spatial and temporal features for accurate class filtration. The structure of our proposed PATAN is as shown in Figure~\ref{figure:s2-3-patan}. In this section, we further describe the implementation of PATAN in detail.

Our networks and experiments are implemented using the PyTorch~\cite{paszke2019pytorch} library. To obtain video features, we instantiate Temporal Relation Network~\cite{zhou2018temporal} as the backbone for video feature extraction for both source domain videos and target domain videos, with the model pretrained on ImageNet~\cite{deng2009imagenet}. The source and target feature extractors share parameters. New layers are trained from scratch, and their learning rates are set to be 10 times that of the pretrained-loaded layers.

The stochastic gradient descent algorithm~\cite{bottou2010large} is used for optimization, with the weight decay set to 0.0001 and the momentum to 0.9. The batch size is set to 8 per GPU. Our initial learning rate is set to 0.005 and is divided by 10 for two times during the training process. We train our networks with a total of 50 epochs for UCF-HMDB\textsubscript{\textit{partial}} and HMDB-ARID\textsubscript{\textit{partial}}, while for MiniKinetics-UCF we train for 30 epochs. The flip-coefficient of the Gradient Reverse Layer (GRL) is increased gradually from 0 to 1 as in DANN~\cite{ganin2015unsupervised}. All experiments are conducted using two NVIDIA RTX 2080 GPUs. 

\clearpage
{\small
\bibliographystyle{ieee_fullname}
\bibliography{iccv}
}

\end{document}